\documentclass[letterpaper, 10 pt, conference]{ieeeconf}
\IEEEoverridecommandlockouts  
\usepackage{algorithm, algpseudocode}  %algorithmic}
\usepackage{graphics}
\usepackage{amsmath, amssymb, bm} % assumes amsmath package installed
\usepackage{xcolor}
\usepackage[normalem]{ulem}
\usepackage{graphicx}
\usepackage{verbatim}
\usepackage{balance}
\usepackage{hyperref}
\usepackage{booktabs}

\newcommand{\pisup} {\pi_{D}} 
\newcommand{\pirob} {\pi}
\newcommand{\cloningloss} {\mathcal{L}(\pirob(s_t),\pisup(s_t))}
\usepackage{caption}
\usepackage{subcaption}

\usepackage{url}

\usepackage{xspace}

\usepackage[backend=bibtex,
            hyperref=true,
            url=false,
            isbn=false,
            doi=false,
            backref=false,
            style=ieee,
            natbib=true,%compatibility aliases
            mincitenames=1,
            maxcitenames=1,
            citestyle=numeric-comp,
            sorting=nyt,%none
            block=none]{biblatex}

\begin{document}
\title{Learning Switching Criteria for Sim2Real Transfer \\ of Robotic Fabric Manipulation Policies}

\author{Satvik Sharma$^{*1}$,
        Ellen Novoseller$^{*1}$,
        Vainavi Viswanath$^{1}$,
        Zaynah Javed$^{1}$,
        Rishi Parikh$^{1}$, \\
        Ryan Hoque$^{1}$,
        Ashwin Balakrishna$^{1}$,
        Daniel S. Brown$^{1}$,
        Ken Goldberg$^{1}$
        % <-this % stops a space
\thanks{$^{1}$Authors are with the University of California, Berkeley, USA.} 
\thanks{$^{*}$ These authors contributed equally.}}

\maketitle

\begin{abstract}
Simulation-to-reality transfer has emerged as a popular and highly successful method to train robotic control policies for a wide variety of tasks. However, it is often challenging to determine when policies trained in simulation are ready to be transferred to the physical world. Deploying policies that have been trained with very little simulation data can result in unreliable and dangerous behaviors on physical hardware. On the other hand, excessive training in simulation can cause policies to overfit to the visual appearance and dynamics of the simulator. In this work, we study strategies to automatically determine when policies trained in simulation can be reliably transferred to a physical robot. We specifically study these ideas in the context of robotic fabric manipulation, in which successful sim2real transfer is especially challenging due to the difficulties of precisely modeling the dynamics and visual appearance of fabric. Results in a fabric smoothing task suggest that our switching criteria correlate well with performance in real. In particular, our confidence-based switching criteria achieve average final fabric coverage of 87.2-93.7\% within 55-60\% of the total training budget. See \href{https://tinyurl.com/lsc-case}{https://tinyurl.com/lsc-case} for code and supplemental materials.
\end{abstract}


% \fi
\IEEEpeerreviewmaketitle

\section{Introduction}

% What is the problem
Training robot control policies in simulation for deployment in physical environments and automation tasks has seen increasing success in a broad variety of robotics applications such as robot grasping~\cite{mahler2017dexnet, hand-eye}, deformable manipulation~\cite{hoquevisuospatial, seita2020deep}, and legged locomotion~\cite{rma2021, bipedal2021}. Simulated training enables the collection of diverse experiences for policy learning that may be infeasible to collect on a physical system, creating potential for strong policy generalization and robustness. Additionally, learning policies purely in simulation can limit the risk of encountering suboptimal behaviors on physical robots and can reduce hardware wear-and-tear. However, transferring policies learned in simulation to the real world is challenging, as simulators are imperfect. As a result, learning algorithms must overcome a domain shift~\cite{valassakis2020crossing} between the dynamics and visual appearance of simulations and the physical world. Prior work has addressed a range of methods to enable robust domain adaptation of policies trained in simulation to physical experiments~\cite{Tanwani2020DIRL, James2019SimToReal, rusu2017sim,  zhao2020sim, bipedal2021,hofer2021sim2real}. Less attention has been given to the critical challenge of determining when a policy trained in simulation is ready for physical deployment. While deploying policies trained with too little simulation data can lead to low performance on the physical system, policies that are trained excessively in simulation risk overfitting to simulator artifacts. This motivates principled methods to determine when policies trained in simulation are likely most ready for physical deployment.

\begin{figure}[t!]
    \centering
    \includegraphics[width=\linewidth]{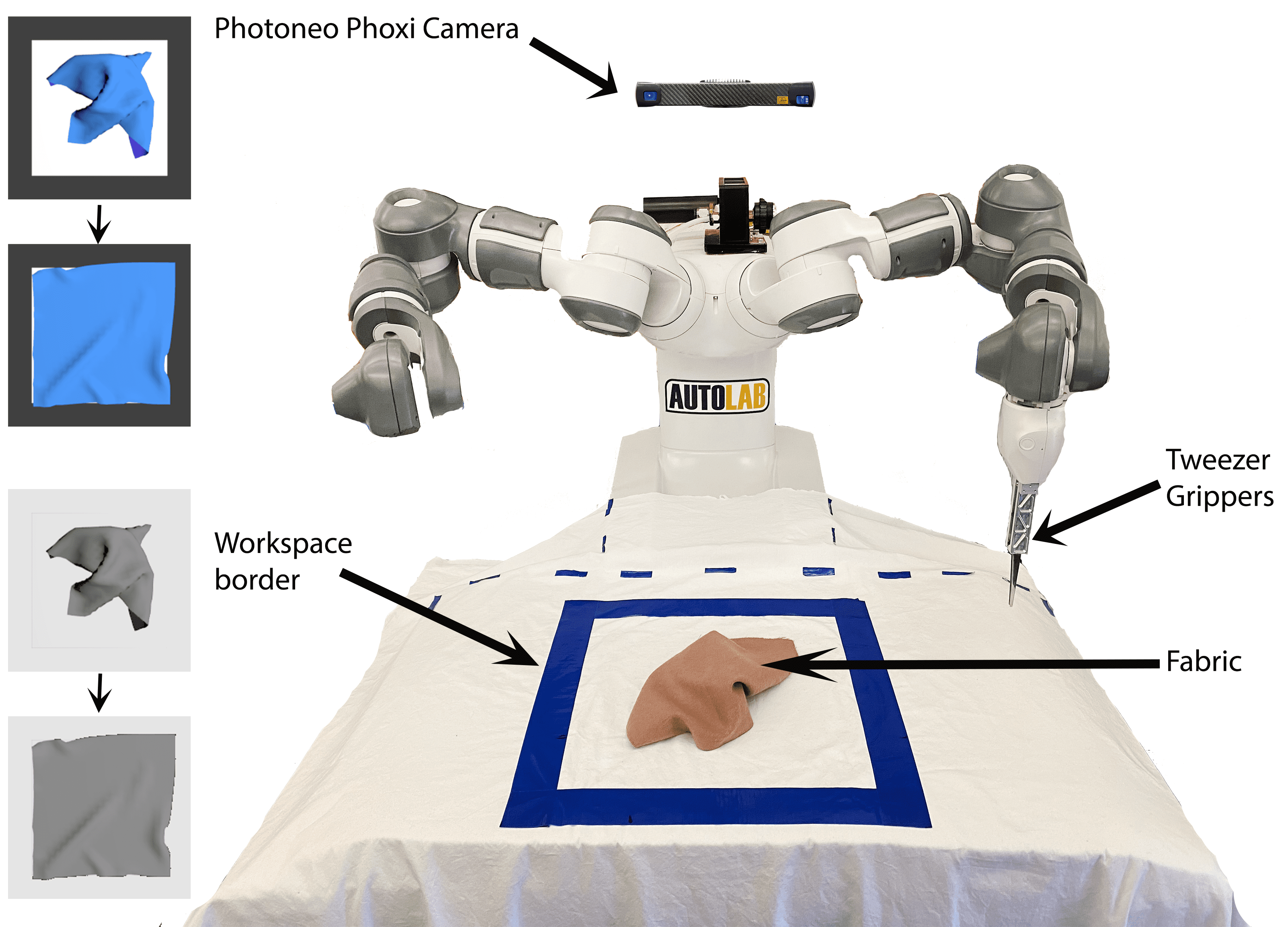}
    \caption{\textbf{Physical experiment and simulator setup.} We study sim-to-real switching in a fabric smoothing task, in an environment consisting of an ABB YuMi robot with a single tweezer gripper. An overhead Photoneo Phoxi Camera captures grayscale images. The manipulation workspace border is marked with blue tape, and the fabric is located within the workspace border. The physical workspace is designed to visually emulate the Gym-Cloth simulator~\cite{seita2020deep}, as shown on the left; the top two lefthand images show example starting and ending configurations from an oracle smoothing policy, while the bottom two images show the same observations, processed to resemble the grayscale images taken by the Photoneo Phoxi Camera.}
    \label{fig:splash}
\end{figure}

% Why is this problem hard/non-trivial
Determining when a policy is ready for physical deployment requires reasoning about its expected performance in physical trials when the policy has only been trained in simulation. One common way to achieve this is to periodically evaluate learned policies in the physical world during simulated training to determine when sufficient performance has been achieved~\cite{DARC, workflow-offline-rl, what-matters-lfd}. However, this poses a number of practical challenges, as performing physical rollouts during training costs significant time and engineering effort, and furthermore may pose safety risks. This motivates using policy rollouts only from simulation to determine when to deploy policies on a physical robot. While this setting is more practical, it is also significantly more difficult, since it is challenging to gain information about expected performance in physical trials at deployment time without access to the physical system prior to deployment.

% \begin{figure}[t!]
%     \centering
%     \includegraphics[height=5cm]{example-image-a}
%     \caption{Placeholder Splash Figure}
%     \label{fig:splash}
% \end{figure}

% What is our insight. 
Our key insight is that the problem of determining when to transfer robotic control policies from simulation to reality bears similarities to the early stopping problem, a well-known technique for reducing overfitting in machine learning \cite{PRECHELT1998761, Mahsereci2017EarlySW, Yao2007}. For instance, \citet{PRECHELT1998761} investigates 14 different early stopping criteria with cross validation, which assesses generalization by measuring performance on a validation dataset unseen by the learner. Meanwhile, other approaches do not rely on the existence of a validation set: \citet{Mahsereci2017EarlySW} derive an early stopping rule based on local gradient statistics and \citet{Yao2007} propose a rule based on the bias-variance tradeoff. While early stopping is well-studied, however, few have explored applying the technique to sim2real transfer. One such work is \citet{Muratore2021AssessingTF}, who propose early stopping in simulation based on an upper confidence bound on the optimality gap during domain randomization.

This work, meanwhile, studies four different mechanisms for switching from simulation to physical experiments and evaluates the degree to which they are predictive of performance during physical deployment.
%in statistical machine learning. When training statistical models, a common approach to prevent overfitting is to track metrics such as model uncertainty or performance on a validation set and use these to determine when to halt training. The intuition here is that even if training performance has room for improvement, this improvement may not translate to improved performance on an unseen test set. Analogously, we study how a number of different metrics measuring learning performance transfer across different simulations.
%, and model uncertainty can be used to determine the potential for policies to transfer to physical experiments. 
Specifically, we focus on robot fabric manipulation, a domain that has seen increasing interest~\cite{seita2020deep, lin2022learning, hoquevisuospatial} and applies across many domains such as home robotics, senior care, laundry folding, and manufacturing. In particular, we study how our proposed sim2real switching criteria can be used to determine when to deploy fabric smoothing policies on a physical robotic system.

% Concrete contributions
This paper contributes (1) a detailed empirical study of which metrics are predictive of robust policy transfer from simulation to the physical world and (2) an evaluation of the proposed method for transferring fabric smoothing policies from simulation to deployment on a physical robot. Fabric smoothing experiments suggest that our switching criteria correlate with performance in real, achieving an average final fabric coverage of 88.79\% over all switching conditions. %Furthermore, our confidence-based switching criteria utilize only 55-60\% of the total training budget.
Furthermore, given a training budget of 200 learning iterations, our confidence-based switching criteria utilize only 55-60\% of total training resources.

\section{Related Work}
% \ashwin{TODO: add more relevant RW + reframe to be in line with new method}
% \ashwin{probably want a section on research in early stopping in machine learning in general}
% \ashwin{would be nice to have some discussion on off-policy evaluation as well, since this is relevant also to determining when you can deploy policies in the real world}

\subsection{Sim2Real Transfer for Robot Learning}
There is significant prior work on learning policies in simulation and facilitating transfer to physical experiments. The most common approach is domain randomization~\cite{mahler2017dexnet, tobin2017domain}, which varies dynamical and/or visual properties such as friction, lighting, camera angle, colors, and textures in simulation to enable zero-shot transfer to the physical world. \citet{bipedal2021} apply dynamics randomization according to a curriculum to achieve physical bipedal robot walking. Similarly, Valassakis et al. \cite{valassakis2020crossing} investigate dynamics domain randomization via injecting random forces into simulations. Domain randomization has achieved transfer in a variety of applications such as robot legged locomotion~\cite{bipedal2021}, fabric manipulation~\cite{hoquevisuospatial}, and robotic grasping~\cite{mahler2017dexnet, tobin2017domain}. Recent work has also explored adaptive methods and curricula for domain randomization, such as automatic domain randomization for dexterous manipulation of a Rubik's cube~\cite{akkaya2019solving} and active domain randomization~\cite{Mehta2019ActiveDR}. Other proposed approaches for crossing the reality gap include learning a canonical intermediate representation \cite{James2019SimToReal}, domain adaptation via a small amount of real data \cite{Tanwani2020DIRL}, and ``real2sim2real" tuning of simulation parameters based on real data \cite{Lim2021PlanarRC, Ramos2019BayesSimAD, Du2021AutoTunedST}. Finally, \citet{habitatsim2real20ral} propose a sim2real correlation coefficient for predicting sim2real performance but require real data to compute the coefficient. Despite the significant body of work on simulated training procedures to enable physical deployment, there has been very little research on automatically determining precisely when to stop training in simulation and deploy learned policies in the physical world without any real evaluation. The closest prior work to ours may be \citet{Muratore2021AssessingTF}, who propose early stopping in simulation based on an upper confidence bound on the optimality gap during domain randomization. However, while this work is specific to the reinforcement learning paradigm, we apply our approach to the imitation learning setting, in which algorithms learn from demonstrations. Furthermore, our proposed framework can readily extend to other settings such as reinforcement learning; this could be an interesting direction for future study.

%\subsection{Early Stopping in Statistical Machine Learning}

%Early stopping is a well-known technique for reducing overfitting in machine learning \cite{PRECHELT1998761, Mahsereci2017EarlySW, Yao2007}. Machine learning practitioners typically assess generalization by measuring performance on a validation dataset unseen by the learner; \citet{PRECHELT1998761} investigate 14 different early stopping criteria with cross validation. Other approaches do not rely on the existence of a validation set: \citet{Mahsereci2017EarlySW} derive an early stopping rule based on local gradient statistics and \citet{Yao2007} propose a rule based on the bias-variance tradeoff. While early stopping is well studied, few have explored applying the technique to sim2real transfer. \citet{Muratore2021AssessingTF} propose early stopping in simulation based on an upper confidence bound on the optimality gap during domain randomization. We study four different mechanisms for switching from simulation to physical experiments and evaluate the degree to which they are predictive of performance during physical deployment.

\subsection{Fabric Manipulation}
Fabric manipulation is a challenging open problem for robots, as fabric has an infinite-dimensional configuration space, exhibits complex dynamics, and often self-occludes. There has been significant prior work in leveraging simulators and task demonstrations to learn fabric smoothing and folding policies for zero-shot transfer to the physical world~\cite{seita2020deep, sim2real_deform_2018, jangir2020dynamic, fabric-descriptors, mmgsd}. Other recent works relax the assumption that task demonstrations are available but still leverage simulation to enable zero-shot transfer to real world experiments~\cite{hoquevisuospatial, lin2020softgym, lerrel}. Another approach is to learn fabric manipulation policies entirely from real world interaction~\cite{finn_vf_2017, lee2020learning}. Our work learns
%We build on the first of these approaches~\cite{seita2020deep, sim2real_deform_2018, jangir2020dynamic, fabric-descriptors, mmgsd}, learning 
fabric manipulation policies purely in simulation for deployment in the physical world; however, unlike prior work that performs zero-shot transfer to physical experiments after some fixed amount of simulated training, we study how to automatically determine when to stop training policies in simulation to facilitate robust deployment on the physical robot.
% While these approaches have shown promise, approaches which rely on zero-shot transfer from simulation to reality can struggle to perform precise tasks in physical experiments due to the inherent domain gap between simulation and reality~\cite{hoquevisuospatial}. While approaches which learn entirely from real world data are less susceptible to these challenges, they can often cause significant wear and tear on the robot due to the amount of data needed to learn reliable fabric manipulation policies \ryan{plus difficulty of cloth state resets in real}. \algabbr{} attempts to bridge this gap, and learns fabric manipulation policies from a combination of interaction with a number of different simulators and the real world, which has not yet been explored in prior work. 
\section{Problem Statement}
\label{sec:problem_statement}
This work considers learning a policy $\pi$ for some task given access to only a computational simulation. In particular, we seek to identify the optimal time at which to terminate learning in simulation and deploy the policy in the physical world. We model the task as a Partially Observable Markov Decision Process (POMDP)~\cite{kaelbling1998planning} $\mathcal{M}_\text{sim}^\phi$, parametrized by simulation parameters $\phi$ capturing variables such as visual appearance and simulator dynamics. The objective is to achieve high task performance when this policy is deployed in the physical world, which we model as POMDP $\mathcal{M}_\text{real}$.

More formally, we assume that $\mathcal{M}_\text{sim}^\phi$ and $\mathcal{M}_\text{real}$ have a shared state space $\mathcal{S}$, observation space $\mathcal{O}$, action space $\mathcal{A}$, reward function $R: \mathcal{S} \rightarrow \mathbb{R}$, initial state distribution $\mu$, and episode time horizon $H$, but may have different mappings from states $s \in \mathcal{S}$ to observations $o \in \mathcal{O}$. In this work, we consider grayscale image observations ($\mathcal{O} = \mathbb{R}^{H \times W \times C}$, $C = 1$), but consider settings (as is typical in practice) where observations corresponding to a specific state may have different visual appearances between simulation ($\mathcal{M}_\text{sim}^\phi$) and reality ($\mathcal{M}_\text{real}$). We additionally consider settings in which the state transition dynamics associated with $\mathcal{M}_\text{sim}^\phi$, denoted by $P_\text{sim}: \mathcal{S} \times \mathcal{A} \times \mathcal{S} \rightarrow [0, 1]$, may be different from those associated with $\mathcal{M}_\text{real}$, denoted by $P_\text{real}: \mathcal{S} \times \mathcal{A} \times \mathcal{S} \rightarrow [0, 1]$. This reflects the inability of computational simulations to precisely model the dynamics of the physical world. 

We consider a setup in which robot policy $\pirob: \mathcal{O} \rightarrow \mathcal{A}$ is first learned with some policy search algorithm in simulation, using $T_\text{sim}$ total transitions in simulated environment $\mathcal{M}^\phi_\text{sim}$. During this phase, we aim to optimize the following objective based on the attained rewards in simulation:
\begin{align}
    \label{eq:sim_obj}
    J^{\phi}_\text{sim}(\pirob) = \mathbb{E}_{\pirob , \mathcal{M}^\phi_\text{sim}} \left[\sum_{t=1}^{H} R(s_t)\right],
\end{align}
where the expectation is with respect to observation-action trajectories sampled from policy $\pirob$ in POMDP $\mathcal{M}^\phi_\text{sim}$. The objective in this work is to identify the optimal stopping time $T_\text{sim}$ such that when policy $\pirob$ is evaluated in the physical world, it maximizes the following objective based on the attained rewards in physical trials:
\begin{align}
    \label{eq:real_obj}
    J_\text{real}(\pirob) = \mathbb{E}_{\pirob, \mathcal{M}_\text{real}} \left[\sum_{t=1}^{H} R(s_t)\right],
\end{align}
where analogously, the expectation is with respect to observation-action trajectories sampled from $\pirob$ in $\mathcal{M}_\text{real}$.
\section{Switching Criteria for Imitation Learning}
\label{sec:methods}

In this work, we consider switching from simulation to real in the case of imitation learning, in which an algorithm learns a policy from sequentially-collected demonstrations. We propose four criteria to determine when to switch from simulated training to physical deployment. In Section~\ref{subsec:method_overview}, we discuss a general framework for evaluating switching criteria, and in Section~\ref{subsec:methods_criteria}, we will describe the two evaluation metrics for switching (reward in simulation and epistemic uncertainty) studied in this work. Then, Section~\ref{sec:stopping} will propose two stopping conditions for identifying switching points (absolute threshold and gradient-based). Pairing the two evaluation metrics and two stopping conditions results in four possible switching criteria, each a combination of a particular switching metric and stopping condition. 

\begin{figure*}[t!]
    \vspace{4 pt}
    \centering
    \includegraphics[width=1.0\textwidth]{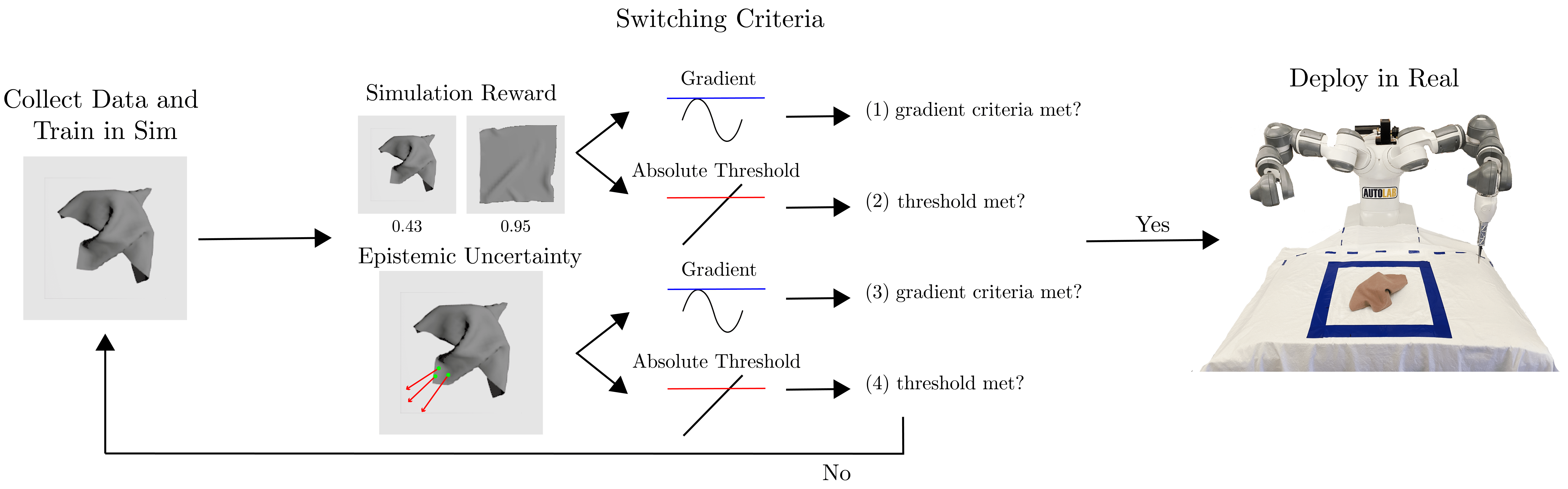}
    \caption{\textbf{System Overview.} At each step, our algorithm pipeline collects a new batch of simulation data, performs a model update epoch, and then checks whether a switching condition is satisfied. If the switching criterion is met, then the model is ready to be deployed in real. Otherwise, we continue collecting simulation data to further update the model. We test four switching criteria, which utilize metrics based on a) reward when evaluated in simulation and b) epistemic uncertainty as estimated via an ensemble of policy networks; these metrics are paired with each of two stopping conditions based on 1) absolute thresholding or values, and 2) gradients.
    %(1) simulation reward, which is the average coverage across 5 iterations in simulation, and the gradient-based stopping condition; (2) the simulation reward and the absolute threshold stopping condition; (3) the epistemic uncertainty, which is the variance in the actions predicted by the ensemble network, and the gradient-based stopping condition; and (4) epistemic uncertainty and the absolute threshold stopping condition.
    }
    \label{fig:alg_overview}
\end{figure*}

\subsection{Switching to Real in Imitation Learning}
\label{subsec:method_overview}
To determine when to switch from learning in simulation to deploying policies in physical experiments, we learn a switching criterion  $\psi: \pirob \rightarrow \{0, 1\}$. If $\psi(\pirob) = 1$, we terminate simulated training and deploy $\pirob$ in physical experiments, while otherwise, we continue training $\pirob$. The switching function $\psi$ identifies a switching time $T_\text{sim}$ as a function of the evolving robot policy $\pirob$. While in this work, we specifically study switching criteria in the context of learning from demonstrations, future work includes exploring similar ideas for other possible policy learning algorithms, including model-based planning with system identification and model-free reinforcement learning.

We consider the imitation learning setting, in which the objective is to learn a robot policy $\pirob$ that emulates task demonstrations collected in simulation from some demonstration policy $\pisup$. In principle, $\pisup$ could be a human controlling the robot in simulation via a teleoperation interface or an algorithmic controller defined using privileged state information present only in simulation; we consider the latter. We train $\pirob$ in simulation with \textit{behavior cloning}~\cite{ross2011reduction, seita2020deep,hoque2021thriftydagger}, which aims to minimize the following loss function to encourage the robot's policy $\pirob: \mathcal{S} \to \mathcal{A}$ to match that of the demonstrator ($\pisup$) on a set of demonstration trajectories $\mathcal{D}$:
\begin{align}
    \label{eq:IL-objective}
    J_{BC}(\pirob) = \mathbb{E}_{(s_t, \pisup(s_t))\sim\mathcal{D}}\left[ \mathcal{L}(\pirob(s_t),\pisup(s_t))\right],
\end{align}
where $\mathcal{L}(\pirob(s),\pisup(s))$ is an action discrepancy measure between $\pirob(s)$ and $\pisup(s)$ (e.g., the squared loss or 0-1 loss; our experiments use the squared loss). 

The key objective in this work is to determine how many training iterations in simulation are required to train $\pirob$ before it should be deployed on a physical system. To this end, we train $\pirob$ to minimize the loss in Equation~\eqref{eq:IL-objective} in simulation on trajectories in $\mathcal{D}$ and compute a stopping criterion $\psi: \pirob \rightarrow \{0, 1\}$. If $\psi(\pirob) = 1$, we terminate simulated training and deploy $\pirob$ in physical experiments. Otherwise, we query $\pisup$ for an additional task demonstration in simulation, aggregate this demonstration into $\mathcal{D}$, and continue. 

The learning process is described in Algorithm~\ref{alg:main} and illustrated in Figure~\ref{fig:alg_overview}. The procedure alternates between 1) collecting a batch of demonstrations from the demonstration policy $\pisup$, 2) appending the demonstrations to the dataset $\mathcal{D}$, 3) performing a model update epoch to minimize the loss function $J_{BC}(\pirob)$ in Equation~\eqref{eq:IL-objective}, and 4) checking whether the switching criterion $\psi(\pirob)$ is satisfied. If $\psi(\pirob) = 1$, then the process is halted and the final learned policy $\pi$ is deployed on the physical robot system. In practice, our experiments collect a batch of $K = 10$ demonstrations per iteration, where we use $K = 10$ to obtain sufficient granularity of information for determining the switching point. Each model update epoch performs 400 gradient steps on the loss function $J_{BC}(\pirob)$ with a minibatch size of 64. We utilize an Adam optimizer with a learning rate of $2.5 * 10^{-4}$ and L2 regularization of $10^{-5}$.

\begin{algorithm}[t]
\caption{Learned Switching Criteria for Sim2Real}
\label{alg:main}
\footnotesize
\begin{algorithmic}[1]
\Require Maximum number of learning iterations $N$; number of demonstrations $K$ collected between each model update and calculation of the stopping criterion $\psi(\pi)$; demonstration policy $\pi_D$; initialized buffer for demonstration data $\mathcal{D} \leftarrow \emptyset$; randomly initialized robot policy $\pirob$
\For{$i \in \{1,\ldots N\}$}
    \State Collect $\tau^{\text{demo}}_j = ((s_t^{(j)}, a_t^{(j)})_{t=1}^{T})$, $j \in \{1, \ldots, K\}$ from $\pisup$
    \State $\mathcal{D} = \mathcal{D} \cup \tau^{\text{demo}}_1 \cup \ldots \cup \tau^{\text{demo}}_K$
    \State $\pirob \leftarrow \arg\min_{\pirob} \mathbb{E}_{(s_t, \pisup(s_t))\sim\mathcal{D}}\left[\cloningloss\right]$
    \If {$\psi(\pi)$}  % \text{ and } $i \% K = 0$
        \State Terminate learning and deploy on physical system
    \EndIf
\EndFor
\end{algorithmic}
\end{algorithm}

\subsection{Switching Metrics}
\label{subsec:methods_criteria}

We propose two switching metrics, which are computed after every model update step in Algorithm~\ref{alg:main}.
%Here, we describe the two switching metrics that we evaluate in this work and provide intuition for why they may indicate the potential of $\pirob$ to transfer well to physical experiments. To this end, we define the following two policy performance metrics below, which are computed after every $K$ demonstrations, which is defined as an evaluation episode. Then for a given switching metric, we initiate a switch when the stopping condition $\psi(\pi)$ (Sec.~\ref{sec:stopping}) is triggered.

\textbf{Simulation Reward:} Here we evaluate $\pirob$ in simulation and compute the average total reward over $L$ rollouts to approximate $J_\text{sim}(\pirob)$ from Equation~\eqref{eq:sim_obj}. These trials also serve as the cross-validation dataset $\mathcal{D}_{\rm cross}$ for tuning our reward-based switching conditions via cross-validation in simulation, as detailed below. In practice, our experiments use $L=5$, which we found to provide sufficiently small standard error. Intuitively, performing well in simulation suggests that the policy may transfer well to real. %As discussed in Sec.~\ref{sec:stopping}, the value and gradient of this quantity can be used to define a stopping condition for training in simulation.

\textbf{Epistemic Uncertainty:} Here we approximate the epistemic uncertainty of $\pirob$ to characterize the policy's confidence in the actions that it predicts. We estimate epistemic uncertainty by training an ensemble of $E$ policies $\{ \pirob^{(i)} \}_{i=1}^{E}$ on bootstrapped minibatches of the training data $\mathcal{D}$. We then estimate the epistemic uncertainty over a holdout demonstration set $\mathcal{D}_{\text{cross}}$, which is not present during model training. (Note that $\mathcal{D}_{\text{cross}}$ has different definitions for the simulation reward and epistemic uncertainty metrics.) Intuitively, the more consistent (low-variance) the policy's predictions on $\mathcal{D}_{\text{cross}}$ across ensemble members, the higher the probability of policy convergence.  Let $\{ a_i(s)\}_{i=1}^{E} \in \mathbb{R}^M$ denote the corresponding actions when each ensemble member is queried at observation $s$. Further, let $a_{ij}(s)$ denote the $j$\textsuperscript{th} component of the action predicted by ensemble member $i$. The epistemic uncertainty is estimated via:
\begin{align}
    \label{eq:epistemic}
    \mathbb{E}_{s \in \mathcal{D}_{\text{cross}}}\left[    \frac{1}{M}\sum_j \text{Var}_i(a_{ij}(s))                 \right],
\end{align}
where $\text{Var}_i(a_{ij}(s))$ denotes the variance over ensemble member $i \in \{1, 2, \hdots E\}$ in action component $j$, and the expectation is taken over observation-action pairs in $\mathcal{D}_{\text{cross}}$.

%\textbf{Sim2Sim Transfer: } Here we consider evaluating $\pirob$ in a number of other simulators with parameters $(\psi_i)_{i=1}^{B}$ and compute the average total reward over $K$ rollouts in each simulator to approximate $J^{\psi_i}_\text{sim}(\pirob)$ for $\pirob$ trained in a simulator with parameter $\phi$ when evaluated in out-of-distribution simulations $(\psi_i)_{i=1}^{B}$.

\subsection{Stopping Conditions}\label{sec:stopping}

We now leverage the two metrics in Section~\ref{subsec:methods_criteria} to propose two stopping conditions for sim2real switching.

\textbf{Value-Based: }
The first stopping condition is an absolute threshold, $A$, which we tune using the cross-validation set,  $\mathcal{D}_{\rm cross}$. For the simulation reward metric, $A$ is set to approximately the maximum reward attained in $\mathcal{D}_{\rm cross}$ (on average over the $L$ rollouts in $\mathcal{D}_{\rm cross}$); notably, $A$ is tuned using $\mathcal{D}_{\rm cross}$, an offline dataset in which rewards plateau to stable performance in simulation, so that future online learning can halt early using $A$. For epistemic uncertainty, $A$ is the highest epistemic uncertainty over $\mathcal{D}_{\rm cross}$ that falls within a preset margin of the minimum epistemic uncertainty value.
%highest epistemic uncertainty that corresponds with the earliest reward peak in $\mathcal{D}_{\rm cross}$. 
Specific values of $A$ are provided in the supplemental website~\cite{supplement}. At each evaluation episode $i$, a spline $f(x)$ is fit to the previous $i-1$ data points. We use a spline---rather than raw data---to mitigate noise in the reward and uncertainty-based switching metrics, as seen in Fig.~\ref{fig:switch}. Then, the spline is evaluated at the current episode and compared with the predefined absolute threshold $A$ to determine whether the algorithm should stop training and deploy the policy in real.

 \textbf{Gradient-Based: }
The second stopping condition is gradient-based, and determines when the change over time in an evaluation metric falls below a threshold. Similarly to the absolute threshold, we identify the threshold by fitting a spline to the previous $i-1$ data points. Then, finite differences are taken for two points evaluated on the spline to approximate the current gradient of the evaluation metric. If the gradient is within the range $[-\epsilon, \epsilon]$ (tuned for each switching metric as described below), the gradient is considered to be sufficiently close to zero and two counters are potentially incremented: $w_{\rm consec}$ keeps track of the number of consecutive episodes with a sufficiently-small gradient, while $w_{\rm total}$ keeps track of the total number of episodes with a below-threshold gradient. The stopping condition is triggered if either $w_{\rm consec} > U$ or $w_{\rm total} > V$, where $U$ is the maximum number of consecutive low-gradient episodes before stopping and $V$ is the maximum total number of (potentially-nonconsecutive) low-gradient episodes. The parameter $V$ allows training to stop even if there are some aberrations that would reset $w_{\rm consec}$ to zero. However, the $V$ condition alone is insufficient, since if the gradient frequently oscillates near zero due to noise in the switching metric, $w_{\rm total}$ would be incremented even though we may not have converged to acceptable performance, in which case the policy can still learn and should not yet be deployed in real. 

Fig~\ref{fig:alg_overview} illustrates this process. In practice, the parameters ($\epsilon, U, V$) of the stopping conditions are tuned via cross-validation in simulation, using the dataset $\mathcal{D}_{\rm cross}$. In particular, $\epsilon$ is empirically set to the largest value such that not many early iterations (which have comparatively larger gradients) are classified as having a below-threshold gradient. $U$ and $V$ are tuned in conjunction so that the stopping condition is triggered as early as possible but not too early (e.g., for an overly-low simulation reward). Specific values for hyperparameters are provided in the supplemental website~\cite{supplement}.

To summarize, our four proposed sim2real switching methods are: 1) simulation reward with the value-based stopping condition, 2) simulation reward with the gradient-based stopping condition, 3) epistemic uncertainty with the value-based stopping condition, and 4) epistemic uncertainty with the gradient-based stopping condition.

% \ashwin{TODO: add more details to above + describe final baselines}

% \daniel{Baselines that come to mind are to train on all sims simulataneously, to just do domain randomization in one sim, and to train for the max sim budget}

\section{Experiments}
The four switching criteria---the two evaluation metrics coupled with each of the two stopping conditions---are each evaluated on a fabric smoothing task, in which we use behavior cloning to learn a policy from demonstrations.

\subsection{Experimental Setup: Fabric Smoothing}
\label{subsec:exp-setup}
% Experimental setup
\subsubsection{Sequential Fabric Smoothing Problem Statement}

We consider the specific manipulation task of sequential fabric smoothing, a challenging open problem in robotics that has received significant recent interest~\cite{seita2020deep, hoquevisuospatial, fabric-descriptors}. As described in prior work~\cite{seita2020deep}, the objective in sequential fabric smoothing is to find a sequence of robot actions to maximally smooth a fabric from an initially crumpled configuration. Concretely, we consider a square crop of fabric with initial configuration (state) $s_0$ and configuration $s_t$ at time $t$. In both simulation and reality, the algorithm can access only top-down grayscale image observations of the workspace, $\mathbf{o}_t \in \mathcal{O} = \mathbb{R}^{H \times W}$, where $H$ and $W$ are the image height and width respectively. Following prior work~\cite{hoque2021lazydagger}, we assume that each side of the fabric is monochromatic and that the two sides are colored differently, where the colors are distinguishable in grayscale (see supplement~\cite{supplement}). At each timestep $t$, the robot executes a 4D pick-and-place action $a_t$ parameterized by $(x_t, y_t, \Delta x_t, \Delta y_t)$, where $(x_t, y_t)$ is the pick point in pixel space and $(x_t + \Delta x_t, y_t + \Delta y_t)$ is the place point. Through a known pixel-to-world transform, the robot picks and lifts the top layer of fabric at $(x_t, y_t)$, translates by $(\Delta x_t, \Delta y_t)$, and releases. The robot seeks to learn a policy $\pi: \mathcal{O} \rightarrow \mathcal{A}$ that maximizes $R(s_T)$, where $T$ denotes the final step of a fabric smoothing episode and $R(\cdot)$ gives the \textit{coverage} or 2D area covered by the fabric (i.e., the area of fabric visible in the workspace). Following prior work~\cite{seita2020deep}, we assume that each smoothing episode terminates when the fabric reaches a coverage threshold of at least 92\% or a limit of 10 actions (whichever happens first).

\begin{figure*}[t!]
    \centering
    \includegraphics[width=0.9\textwidth]{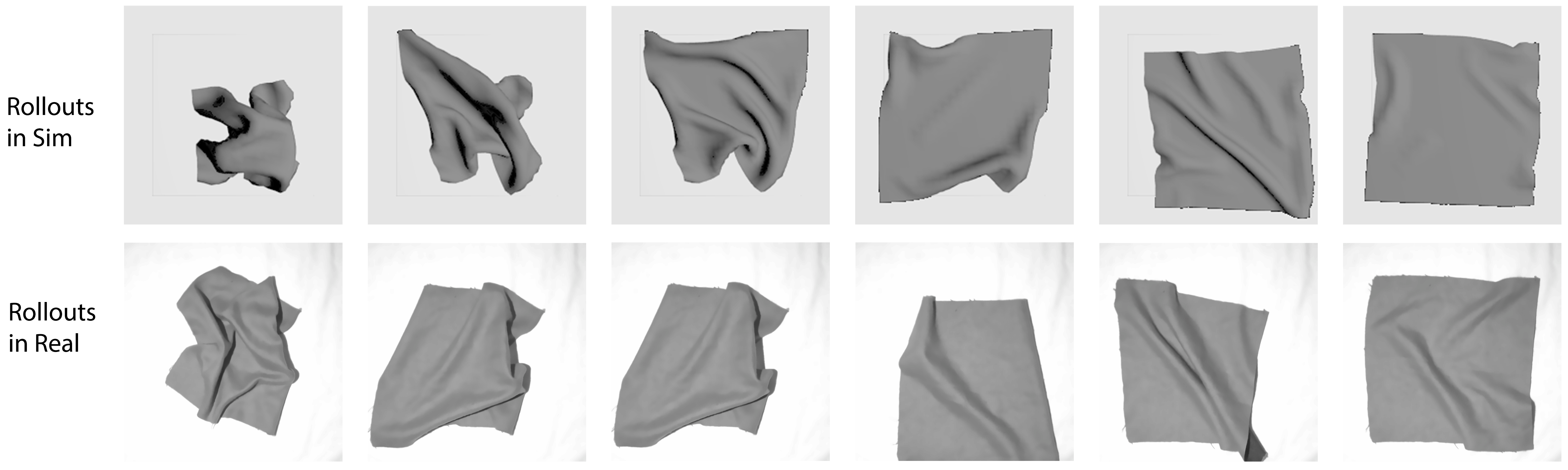}
    \caption{\textbf{Example rollouts in simulation and physical experiments.} The top row (left to right) depicts a sample trajectory in simulation, while the bottom row similarly depicts a sample trajectory in real.}
    \label{fig:iterations}
\end{figure*}

\subsubsection{Fabric Smoothing Simulation Environment}
To collect demonstrations, we use the oracle corner-pulling policy in the Gym-Cloth simulator~\cite{seita2020deep}, which is ideal for fabric manipulation tasks. Gym-Cloth is an OpenAI-Gym-style fabric manipulation environment, which simulates the cloth using structural, shear, and flexion springs. We use this simulator to collect data for training and testing our policies in simulation before switching to deployment in real. The cloth is represented as a 25x25 grid of point masses and is pictured with one light blue side and one dark blue side. The images generated for learning are 224x224 px, of which a completely smooth cloth would occupy 164x164 px. This ratio is reflected in our physical setup. Lastly, when using the Gym-Cloth-simulated data to train fabric smoothing policies, we perform color shifts and augmentations. Specifically, the colors of the background and both sides of the cloth are darkened, converted to gray scale, and blurred by a constant amount to match the images in real. In Fig~\ref{fig:iterations}, rollouts in simulation and real show the simulation images adapted to closely resemble real images. We collect a demonstration dataset $\mathcal{D}$ of 2,000 episodes of the oracle corner-pulling policies in Gym-Cloth for training behavior cloning policies.

\subsubsection{Fabric Smoothing Physical Environment}
The workspace, depicted in Fig.~\ref{fig:splash}, contains a bimanual ABB YuMi robot with a single tweezer gripper on its left arm; we do not use the right arm. Tweezer grippers are ideal for grasping the cloth as they are able to apply fine point pressure, unlike many standard grippers~\cite{seita2020deep}. The manipulation surface is white and foam-padded to avoid end effector damage during any workspace collisions. The manipulation workspace is marked with blue tape (see Fig.~\ref{fig:splash}), so that its size with respect to the cloth size resembles the setup in the Gym-Cloth simulator. We use a double sided 25 cm x 25 cm cloth with one light brown and one dark brown side. The workspace has an overhead PhotoNeo Phoxi Camera that captures grayscale images of resolution 732 x 1142 px, which are later cropped to specifically display the manipulation workspace, enclosed by the blue tape. To ease cloth perception, we also normalize the physical camera images based on a constant affine transformation to create greater contrast between the cloth and background. We further blur the images to remove noise, making them more closely resemble the smooth images seen in the Gym-Cloth simulator.

Our physical experiments evaluate four repetitions of each policy; for repeatability, these utilize a four fixed initial fabric configurations, pictured on the supplemental website~\cite{supplement}. During physical experiments, we project fabric pick points onto a color-segmented mask of the fabric, and project place points within the workspace. Similarly, when evaluating learned policies in the simulator, we project pick points onto a mask of the fabric. Additionally, in simulation, episodes are terminated early if the fabric leaves the workspace.

% Metrics
\subsection{Evaluation Metrics}
\label{subsec:exp-eval}
Our experiments evaluate the following switching criteria introduced in Sec.~\ref{sec:methods} to determine when the behavior cloning policy should stop training and be deployed in real:
\begin{enumerate}
    \item (Reward value) Simulation reward with a value-based stopping condition,
    \item (Reward gradient) Simulation reward with a gradient-based stopping condition,
    \item (Confidence value) Epistemic uncertainty with a value-based stopping condition, and
    \item (Confidence gradient) Epistemic uncertainty with a gradient-based stopping condition.
\end{enumerate} 

%\rishi{We need to label these graphs to make it clear that these are the 4 switching conditions. Use the same exact wording everywhere. Have a legend to explain the stopping conditions. (All four metrics tell us when to stop.)}
%\rishi{Why do we see differences in table one if they all stabilize around the same time? The differences in table one could all be due to noise?}
%\rishi{We see platues around the same area}

\begin{figure*}[t!]
    \vspace{4 pt}
    \centering
    \includegraphics[width=0.24\textwidth]{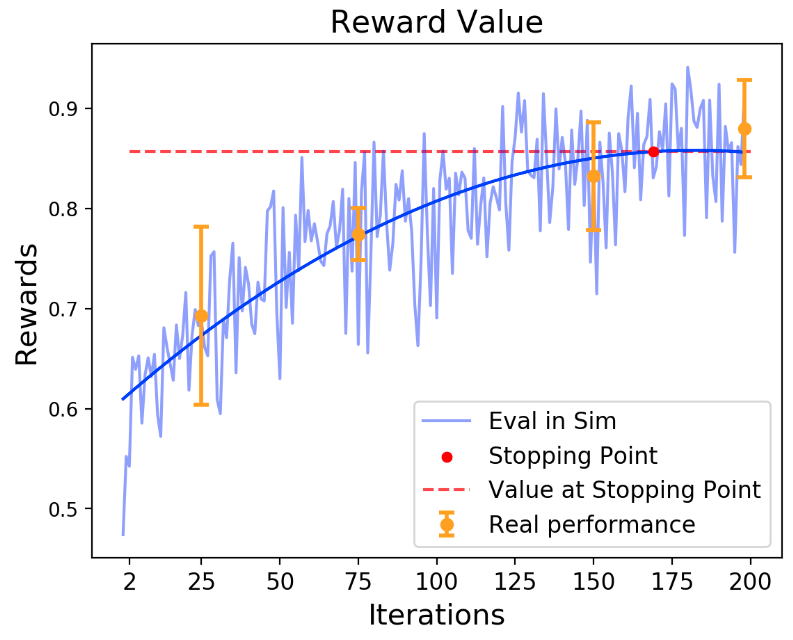}
    \includegraphics[width=0.24\textwidth]{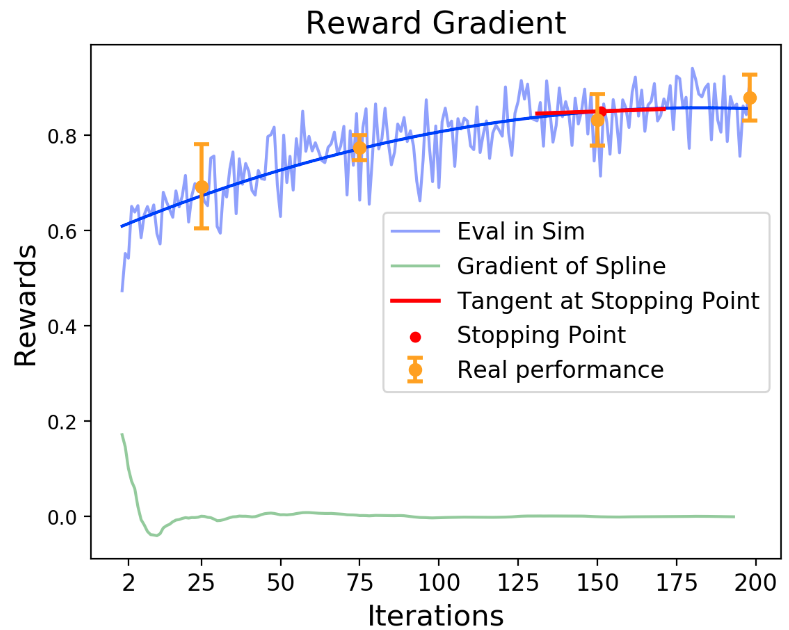}
    \includegraphics[width=0.24\textwidth]{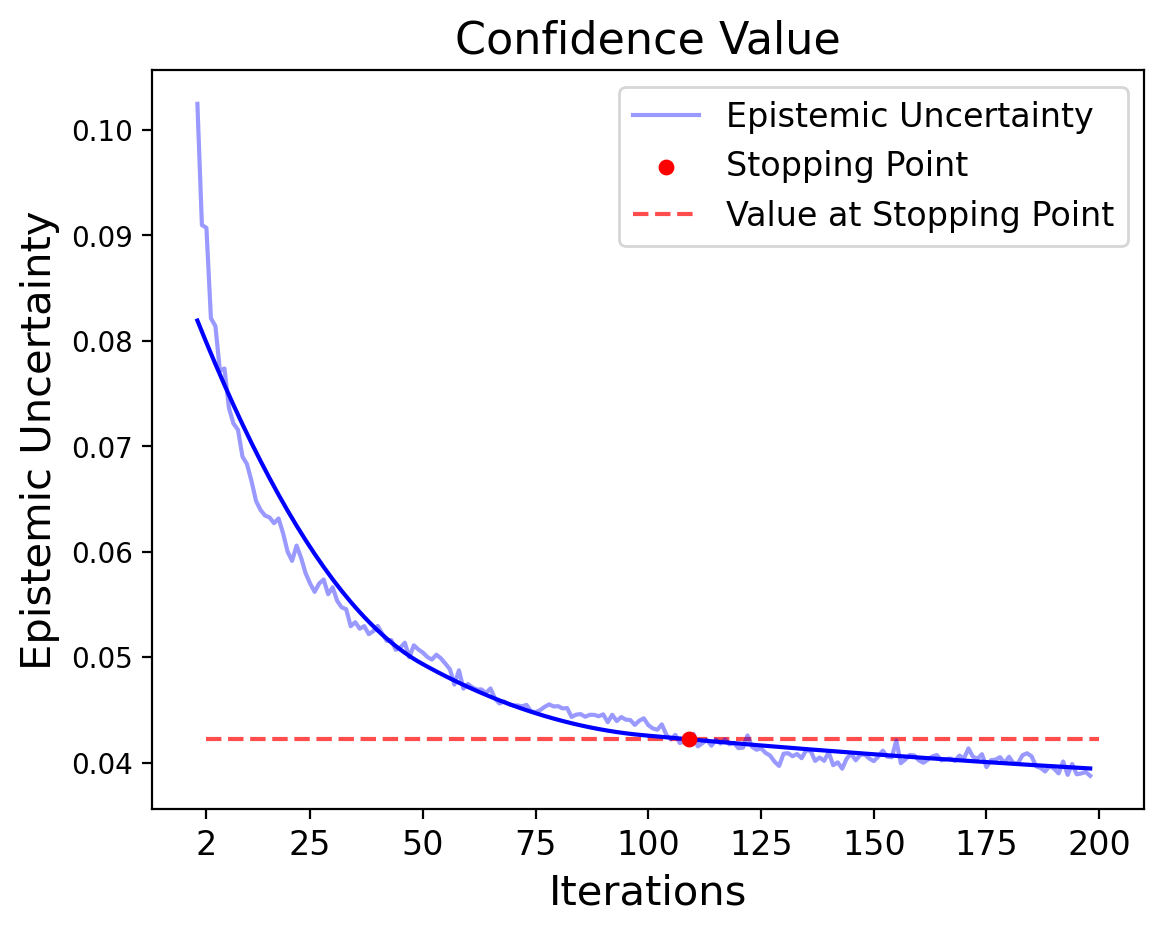}
    \includegraphics[width=0.24\textwidth]{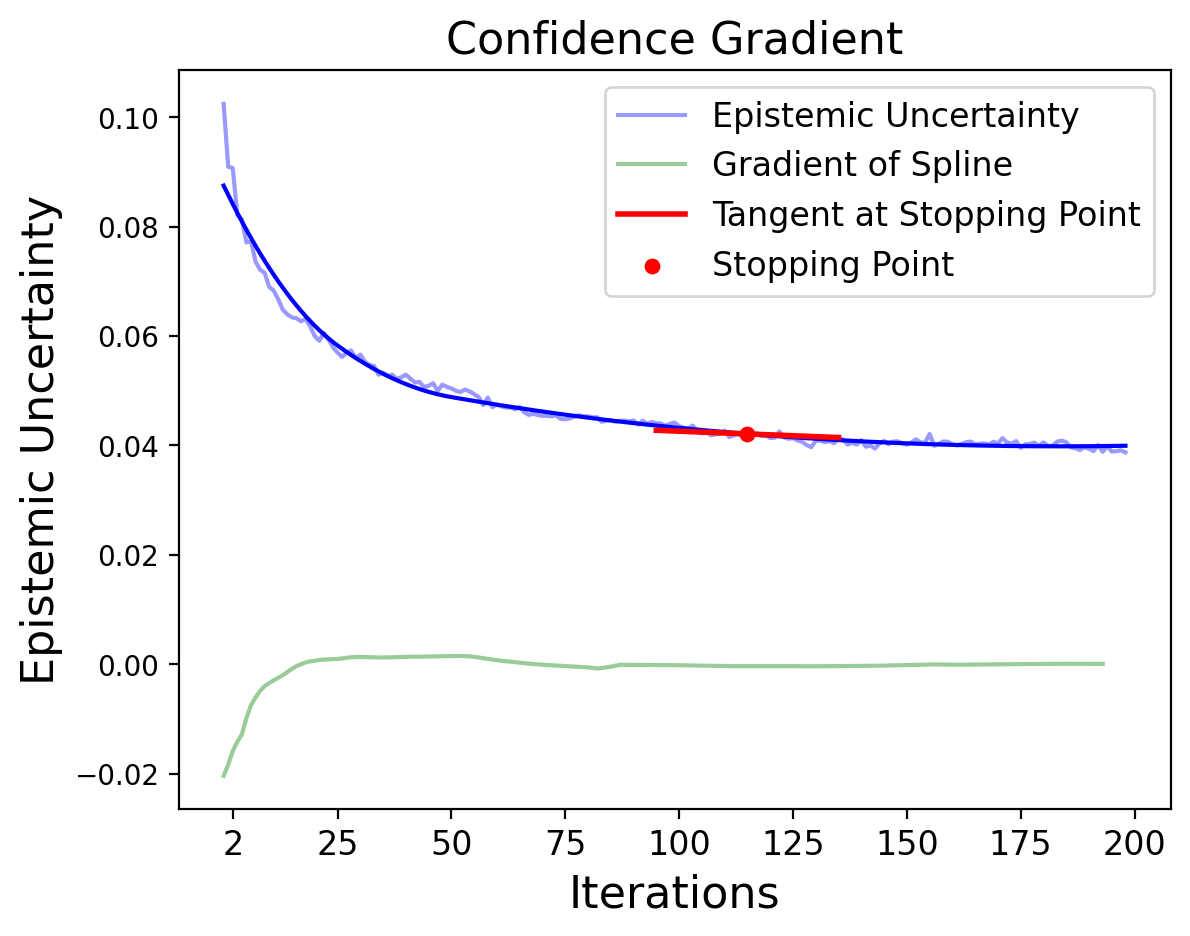}
    \caption{\textbf{Determining stopping points for various switching criteria.} On all graphs, the dark blue curves are splines fit to the data to mitigate noise when approximating the gradient and determine if the value-based threshold has been met. \textbf{Left two graphs:} The simulation reward comes from evaluating the policy in the Gym-Cloth simulation environment and determining the final fabric coverage in each episode; curves are averaged over 5 episode rollouts in Gym-Cloth. For comparison with real, the orange points correspond to mean performance in real of the behavior cloning policy selected at that iteration. The error bars correspond to the standard error across four runs. The stopping points (red point) are determined to be at iteration 171 for reward value and 153 for the reward gradient.
    \textbf{Right two graphs}:
    The epistemic uncertainty is calculated at each iteration over a holdout set of 200 demonstration episodes and with five ensemble members. The confidence value determines the stopping point to be 111, while the confidence gradient determines it to be 117.}
    \label{fig:switch}
\end{figure*}

We evaluate the performance of these switching criteria via the following metrics, calculated during deployment in physical experiments after switching (averaged over 4 episodes in each case):
\begin{enumerate}
    \item Final coverage
    \item Improvement ratio (final coverage / initial coverage)
    \item Number of actions per episode.
\end{enumerate}

%\textbf{TODO:}Basically describe reward + success rate in physical trials after having switched as the evaluation metric

% Results
\subsection{Experimental Results}
\label{subsec:exp-results}
% Show results

\begin{figure*}[t!]
    \vspace{4 pt}
    \centering
    \includegraphics[width=0.39\textwidth]{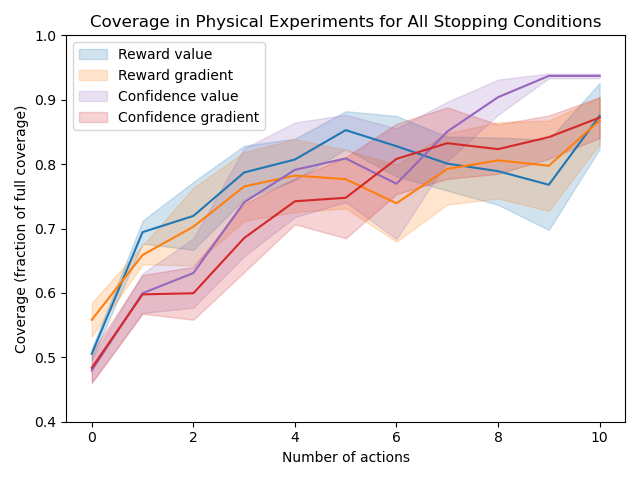}\hspace{10mm}
    \includegraphics[width=0.51\textwidth]{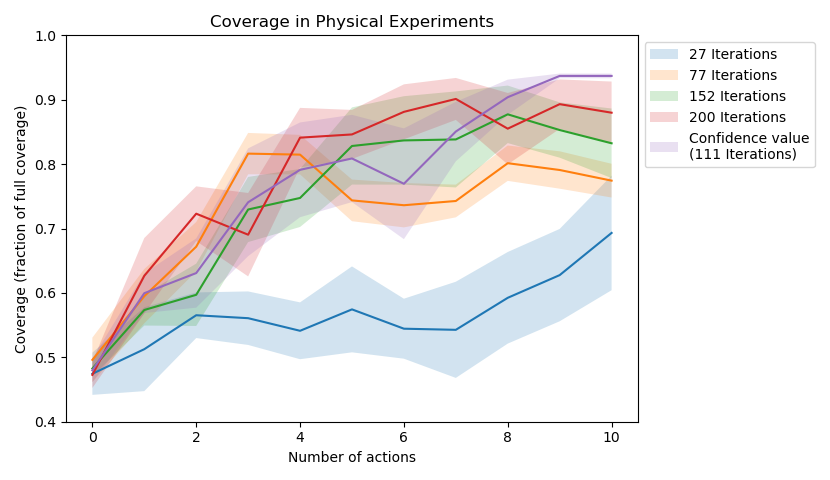}
    \caption{\textbf{Performance of learned policies in physical fabric smoothing experiments at various stopping points.} Left: Final physical fabric coverage achieved for each of the four stopping conditions. Right: Comparing the final physical fabric coverage for the confidence value stopping condition to various checkpoints. We see that the stopping conditions are largely competitive with 200 iterations (the maximum iteration number considered), but require significantly less training. Plots show mean $\pm$ standard error over 4 episodes. Note that to plot episodes that reach the target coverage of 92\% in fewer than 10 actions, we repeat the final achieved coverage for the remainder of the 10-action budget. %\daniel{not sure what our main point here is? It seems to show we should just train as long as possible, but there are some interesting dips in the right plot. Is the idea to show that real performance isn't always clear from sim perf?}
    }
    \label{fig:coverage}
\end{figure*}

\begin{table}[!t]
\centering
\begin{tabular}{ l  c  c  c  c} 
\toprule
Method (iters) & Imp. ratio & Final & Actions \\ 
\midrule
Rew val (171) & 1.732 $\pm$ 0.104 & 0.875 $\pm$ 0.051 & 8.75 $\pm$ 1.08 \\  % 2.598
Rew grad (153) & 1.568 $\pm$ 0.102 & 0.868 $\pm$ 0.037 & 8.25 $\pm$ 1.24\\ % 2.165
Conf val (111) &  1.964 $\pm$ 0.070 & 0.937 $\pm$ 0.004 & 6.00 $\pm$ 1.27 \\ % 2.550
Conf grad (117) &  1.819 $\pm$ 0.103 & 0.872 $\pm$ 0.032 & 8.75 $\pm$ 1.08 \\ % 2.165
Final (200) &  1.858 $\pm$ 0.040 & 0.880 $\pm$ 0.048 & 7.50 $\pm$ 1.30 \\  % 2.384
\bottomrule
\end{tabular}
\\
\caption{\textbf{Physical experiment results.} We report the final fabric coverage and improvement ratio (final / initial coverage) achieved in physical fabric smoothing experiments by the various switching policies: reward value (rew val), reward gradient (rew grad), confidence value (conf val), and confidence gradient (conf grad). We also compare with the performance after 200 learning iterations, the maximum training budget considered. Results are mean $\pm$ standard error over 4 episodes in each case, and the initial coverage averaged 0.500 $\pm$ 0.026 standard error over all cases reported in the table. We also report the average number of actions (where each trajectory terminates upon reaching 10 actions or 92\% coverage, whichever occurs first). %\daniel{Why report start and final? I think it would be better to report improvement over starting coverage so you just have one number that is easier to interpret. }
}
\label{table:phys_exp_results}
\vspace{-2mm}
\end{table}

Fig.~\ref{fig:switch} illustrates the switching calculation, displaying the evaluation metrics and stopping point for each method. Out of a maximum of 200 learning iterations, the stopping points are 111 and 117 for the confidence-based metrics, while they are 171 and 153 for the reward-based metrics. As seen in the figure, the confidence-based switching points occur after fewer learning iterations because the epistemic uncertainty plateaus earlier than the average simulation reward.
%In each case, the stopping point is between 100 and 125 iterations out of a maximum of 200 learning iterations; the values are relatively close because all stopping condition parameters were tuned via cross-validation in simulation.

Table~\ref{table:phys_exp_results} summarizes the experimental results. We see that the average final coverage achieved across all four stopping criteria is 88.79\%, which exceeds the average initial coverage by a factor of 1.78. Thus, all four switching conditions are well-correlated with performance in real. Plotting fabric coverage over time in Figure~\ref{fig:coverage}, we also see that 200 learning iterations---the maximum number considered---are not necessary to achieve competitive performance. Rather, our switching criteria identify earlier stopping times that yield similar final coverage to the 200-iteration comparison, saving training time.

Comparing results across the switching criteria, we see that the confidence-based metric yields significantly earlier stopping points than the simulation reward metric, which plateaus more slowly. Yet, the reward-based switching points do not result in improved physical fabric coverage relative to the confidence-based switching points. Thus, it may be that the simulation reward metric continues to fit to the simulator beyond the point where such fitting benefits physical task performance.
We also see that the epistemic uncertainty metric yields higher final coverage than the simulation reward for both the value-based and gradient-based stopping conditions. Similarly, the value-based stopping condition shows higher performance than the gradient stopping condition for both the epistemic uncertainty and simulation reward metrics. However,
%because the stopping condition iterations are close together (see Table~\ref{table:phys_exp_results}), 
it is possible that these discrepancies partially reflect noise in the behavior cloning training process, rather than an inherent difference between the switching criteria.

Further comparing the methods, Figure~\ref{fig:switch} suggests that the epistemic uncertainty exhibits less iteration-to-iteration noise than the simulation reward, while the simulation reward is easier to tune and interpret and only requires training a single behavior cloning policy (rather than a policy ensemble). Meanwhile, the value-based stopping condition is easier to tune than the gradient-based method. However, because the gradient-based stopping condition uses a time-based trend rather than individual values, it may yield more stable values and improved generalization compared to the value-based stopping condition. Testing generalizability of the stopping conditions is an interesting direction for future work.

%Therefore, while we do see variations in performance on different stopping conditions in Table ~\ref{table:phys_exp_results}, this is not necessarily because one stopping condition yields better results. However, all four stopping conditions together do identify a point as early as possible in training that will achieve close to the maximum possible coverage in the cloth smoothing task. We can make this claim by observing Fig.~\ref{fig:exp_results}. Experiments trained with 150 and 198 iterations converge to around 0.8-0.9 final coverage. Our switching criteria are able to find an earlier time in iterations that will still result in optimal performance with respect to final coverage. Examples of simulation and physical experiment rollouts are shown in Fig.~\ref{fig:coverage}. 

\section{Discussion and Future Work}

This work proposes and evaluates strategies for determining when to switch from training a learning algorithm in simulation to deploying it on a physical robot. We consider metrics based on simulation performance and epistemic uncertainty and suggest stopping conditions based on absolute thresholding and gradient tracking. We apply the method to sim2real transfer for behavior cloning in a fabric smoothing task. Results suggest that our switching criteria correspond well with performance in real and that the proposed switching criteria can save training time by early stopping, which can save compute resources and help to deploy robots in the real world faster. There are many directions for future work. In particular, we will study how the switching criteria generalize across multiple simulators and tasks. We will also consider switching to real based on policy performance when transferring across simulators. Finally, we are also excited about switching to real in other algorithmic contexts, such as model-based planning and model-free reinforcement learning, as well as two-way switching between learning in simulation and in physical experiments.

%\daniel{Seems like all the switching methods work pretty similarly. I think we can sell this as a good thing since it shows that simple methods can work and it seems like we have some convincing results showing that we can save training time by early stopping which is important since it saves compute and energy resources and can get robots out in the real world faster which seems important for industrial applications.}

%\daniel{One thing I've noticed with reviewers is that they like to see a clear winner in the results so at least in the discussion we should say what we think the best switching method is. Probably the one that has the highest coverage improvement and a low number of actions and is also the easiest to implement in practice.}


\section*{Acknowledgment}

The authors would like to thank Mike Danielczuk and Justin Kerr for helpful advice. This research was performed at the AUTOLAB at UC Berkeley in affiliation with the Berkeley AI Research (BAIR) Lab, the CITRIS “People and Robots” (CPAR) Initiative, and the Real-Time Intelligent Secure Execution (RISE) Lab. The authors were supported a CRA Computing Innovation Fellowship and by equipment grants from PhotoNeo, NVidia, and Intuitive Surgical.



% % REPLACE WITH:
% \bibliographystyle{IEEEtran}
% %\balance
% \bibliography{References}

\renewcommand*{\bibfont}{\footnotesize}
\printbibliography %for biblatex -- doesnt work with natbib

\end{document}